\documentclass[10pt, english,pra,aps,twocolumn,floatfix,superscriptaddress,nofootinbib]{revtex4-2}
\usepackage{comment}
\usepackage{braket}
\usepackage{amsmath}
\usepackage{float}
\usepackage{color, colortbl}
\usepackage{tikz}
\usetikzlibrary{quantikz}
\usetikzlibrary{positioning,shadows,shapes,arrows}
\usepackage{dsfont}
\usepackage{amsfonts}
\usepackage{geometry}
\usepackage{xcolor}
\usepackage{graphicx}
\usepackage{dcolumn}
\usepackage{bm}
\usepackage{hyperref}
\usepackage{microtype}
\usetikzlibrary{arrows}
\usetikzlibrary{arrows.meta}
\usetikzlibrary{shapes,arrows}
\usepackage{amsmath,bm,times}
\usepackage{tikz}
\usetikzlibrary{arrows.meta}
\usepackage[english]{babel}

\definecolor{Gray}{gray}{0.9}
\usepackage[english]{babel}
\usepackage{amsthm}

\newcommand{\targettweet}{target tweet}
\newcommand{\fakenews}{fake}
\newcommand{\truenews}{real}
\newcommand{\clustering}{Clustering model}
\newcommand{\crosscheck}{Cross-checking model}
\newcommand{\reliable}{reliable sources}

\newcommand{\rom}[1]{\uppercase\expandafter{\romannumeral #1\relax}}

\begin{document}

\title{Automated Fake News Detection using cross-checking with reliable sources} 
\author{Zahra Ghadiri}
\author{Milad Ranjbar}
\author{Fakhteh Ghanbarnejad}
\email{fakhteh.ghanbarnejad@gmail.com}
\author{Sadegh Raeisi}
\email{sraeisi@sharif.edu}
\affiliation{Department of Physics, Sharif University of Technology, Tehran, Iran}

\begin{abstract}
Over the past decade, fake news and misinformation have turned into a major problem that has impacted different aspects of our lives, including politics and public health. 
Inspired by natural human behavior, we present an approach that automates the detection of fake news. Natural human behavior is to cross-check new information with reliable sources.
We use Natural Language Processing (NLP) and build a machine learning (ML) model that automates the process of cross-checking new information with a set of predefined reliable sources. We implement this for Twitter and build a model that flags fake tweets. Specifically, for a given tweet, we use its text to find relevant news from reliable news agencies. We then train a Random Forest model that checks if the textual content of the tweet is aligned with the trusted news. If it is not, the tweet is classified as fake. This approach can be generally applied to any kind of information and is not limited to a specific news story or a category of information. Our implementation of this approach gives a $70\%$ accuracy which outperforms other generic fake-news classification models. 
These results pave the way towards a more sensible and natural approach to fake news detection.
\end{abstract}
\maketitle

\section{Introduction}
Today social media has become popular among people worldwide as their primary source of news. For example, about half of U.S. adults get news from social media \cite{shearer2021news}. 
Moreover, online social media is becoming the main platform for people to exchange information, opinions, and experiences about different events \cite{matsa2018news}. 
These platforms have significantly helped the free speech, political and social awareness (e.g., the \#metoo movement \cite{xiong2019hashtag}), and raising the voice of minorities and oppressed groups  (e.g., the Black Lives Matter movement) \cite{reuter2015online}. Despite all the benefits, social media can be harmful as well. One of the harms of social media is the spread of misinformation.\\ 
Misinformation and its propagation are not limited to the modern information age and social media. 
A famous example is the misinformation campaign launched in ancient Rome by Caesar Augustus (Octavian), the first Roman emperor, against his opposition, Roman politician, and general, Mark Antony. Specifically, Octavian used slogans etched onto coins to smear the reputation of Antony \cite{kaminska2017module}.\\
In the information age, the lack of proper monitoring and fact-checking and capability of the spread of news by bots \cite{shao2017spread} have turned social media into a key conduit for \fakenews{} news. \\
The spread of misinformation on these platforms can significantly affect different aspects of our lives. 
For instance, some believe that the spread of targeted misinformation has contributed to the 2016 U.S. presidential election \cite{allcott2017social}. 
Some research also indicates that exposure to \fakenews{} news can cause attitudes of inefficacy and cynicism toward certain political candidates \cite{balmas,allcott2017social}. The influence of misinformation is not limited to Politics. Another study shows that 1225 fake news stories were spread during COVID-19, with half of them coming from social media, putting public health in danger \cite{naeem2021exploration}.
\\
This significant effect of misinformation on society has motivated extensive research on \fakenews{} news, especially on Twitter as a major social media. Some of these studies focus on the statistical behavior of \fakenews{} news. A recent study by Vosoughi et al. shows that diffusion parameters of \fakenews{} news are different from \truenews{} news on Twitter. They showed that ``falsehood'' diffuses deeper and faster \cite{vosoughi2018spread}. In another study, \fakenews{} and \truenews{} news propagate different diffusion topologies on social platforms such as Weibo and Twitter \cite{zhao2020fake}.\\
In parallel, some efforts focus on the detection of \fakenews{} news. There were about 137 active fact-checking projects in the world in 2018 \cite{pavleska2018performance}. 
For instance, there are platforms such as PolitiFact \footnote{https://www.politifact.com/}, which uses editors to determine news veracity. Unfortunately, manual fact-checking and \fakenews{} news detection is no match for a large amount of misinformation that is spread on a daily basis, and despite all these efforts in detecting \fakenews{} news, there is a lot of \fakenews{} news which is not detected.
Automatic \fakenews{} news detection provides an effective alternative solution that reduces the human cost for detection and prevents the spread of \fakenews{} news.\\
Various models have been developed to automate the process of \fakenews{} news detection. Each of these models can accurately detect \fakenews{} news among a specific tweet category. We define this characteristic of models as the \emph{effective zone} of the model. An effective zone might contain political tweets, tweets about a famous hurricane, or tweets by Twitter bots. Also, each model uses features such as the text of the tweet, the user who tweets it, tweet characteristics (e.g., number of retweets), and diffusion topologies to detect whether it is \fakenews{} or not. We call this characteristic of a model the \emph{features} of the model. Finally, the \emph{ensemble} is the data that we use to train and test our model, which can characterize the effective zone of the model. Table \ref{table:others_examples} shows some examples of such studies.

\begin{table*}
\centering
{
\renewcommand{\arraystretch}{1.0}
\begin{tabular}
{|>{\centering\arraybackslash}m{.15\linewidth}|>{\centering\arraybackslash}m{.3\linewidth}|>{\centering\arraybackslash}m{.15\linewidth}|>{\centering\arraybackslash}m{.4\linewidth}|}
\hline
\multicolumn{4}{|c|}{Related Work Description} \\ 
\hline \centering
Article& Ensemble & Features & Effective Zone \\
\hline \centering
 Kwon et al. & 102 news topics, each containing at least 60 tweets and their diffusion networks&  Characteristics of diffusion network& The collection of tweets that the diffusion network is available for them\\
\hline \centering
 Helmstetter et al. &  Tweets from a trustworthy source labeled as real news, and tweets from an untrustworthy source labeled as fake news &  User, text, and tweet features & Fake news tweeted from users who almost always tweet fake\\
\hline \centering
Buntain et al. & CREDBANK and PHEME datasets & Tweet, characteristics of diffusion network, user, text features & when we have the structure of tweets, many tweets with the same topic, and users who spread only fake or real most of the time.\\
\hline \centering
Agarwal et al.,2019 & 12.8 K manually labeled short statements & Text features & Since most of the tweets are political, works better in political tweets, but otherwise can be used for other ensembles
\\ \hline \rowcolor{lightgray} \centering
Our Approach & 4k random tweets & Text features & Topics which are mentioned in reliable sources
\\ \hline
\end{tabular}
}
\caption{Related works comparison: This table shows the kind of tweets different papers have focused on their verification and features they used to study them. }
\label{table:others_examples}
\end{table*}

Papers such as Kwon et al \cite{kwon2013prominent} study diffusion network of tweets in order to identify rumors. They use temporal, structural, and linguistic characteristics of diffusion as features of the model on 102 news topics, each containing at least 60 tweets as their ensemble. This model's effective zone is the collection of tweets that the diffusion network is available for them. An issue for using network features in detecting \fakenews{} news is that repeating false information, even in a fact-checking context, may increase an individual's likelihood of accepting it as true \cite{ecker2017reminders}. 
Therefore, it is crucial to detect \fakenews{} news as soon as possible, which is not always possible in studying the diffusion network of news.\\
Another feature to use in studying \fakenews{} news is the information we have on the user who tweeted it, the number of followers, favorites, and Twitter handle, to mention a few. For instance, \cite{helmstetter2018weakly} ensemble contains tweets from a trustworthy source labeled as \truenews{} news and tweet from an untrustworthy source labeled as \fakenews{} news. This model uses features such as user, tweet, topic, and sentiment, and its effective zone is detecting \fakenews{} news tweeted from users who almost always tweet \fakenews{}, such as rumor bots. As another example, Buntain et al. \cite{buntain2017automatically} uses structural, user, and content features of tweets to detect \fakenews{} news. In order for this model to be effective, we should have the structure of tweets, many tweets with the same topic, and users should spread only \fakenews{} or \truenews{} most of the time. Using user features, although applicable to detect \fakenews{} news bots, might lead to a bias through certain users and disables us to predict \fakenews{} news among the tweets of users who often tweet \truenews{} news.\\
Among other features, extracting features from news' text using NLP methods can help us find a solution for the \fakenews{} news detection problem. Many previous studies on tweet texts focus on detecting the veracity of tweets on popular threads on Twitter \cite{qazvinian2011rumor}. These models benefit from an abundance of labeled tweets (as \fakenews{} or \truenews{}) in the case of popular threads. Despite their good results in the case of popular threads, these models might not be as useful in the case of news content that has not been tweeted as much. Most relevant to our study is the 2019 VasuAgarwal et al. \cite{agarwal2019analysis}. This study uses a set of political tweets as the ensemble to train a \fakenews{} news detector based on features extracted from the tweet's text. This set is diverse in terms of political news content, which makes this model effective for detecting \fakenews{} news among both popular and unpopular political threads.\\
In this paper, we present a method for automating \fakenews{} news detection on Twitter. Our approach leads to a model which can also predict the veracity of unpopular threads in any given subject accurately. 
Our approach is based on human behavior when faced with new information or news.
People usually search for reliable news agencies or fact-checking websites to cross-check news and to see if it is \truenews{} or not.\\
Our idea is to automate this process using NLP techniques to build a machine learning model. The objective of the model is to cross-check the content of a given tweet with \reliable{}   and make a prediction about the veracity of the tweet. 
Thus, in addition, we need to prepare a group of \reliable{} in order to cross-check the tweets with it. In other words, our dataset consists of two main data; one is \targettweet{}s, which should be labeled for the machine learning part, and the other part contains reliable tweets that are needed in cross-checking \targettweet{} with them. We select our \reliable{} from news agencies located in the green zone of the “media bias chart”\footnote{https://www.adfontesmedia.com/static-mbc/}. This includes Reuters, BBC, AP, etc. The green box in this chart shows the most reliable  news agencies. In order to compare a \targettweet{} with trustable sources, we collect tweets from the chosen reliable sources. We collect tweets published in the same time interval as the \targettweet{}. We call this dataset the reliable dataset.\\
This paper is organized as follows. The section Methods presents the model, the feature extraction techniques, and the entire pipeline. In the Results, we evaluate the performance of our model. The Discussion and Conclusion section summarizes our work and discusses some aspects of the work that can be improved in future studies.

\section{Method and Data}

\begin{figure*}
\centering
\includegraphics[scale=0.34]{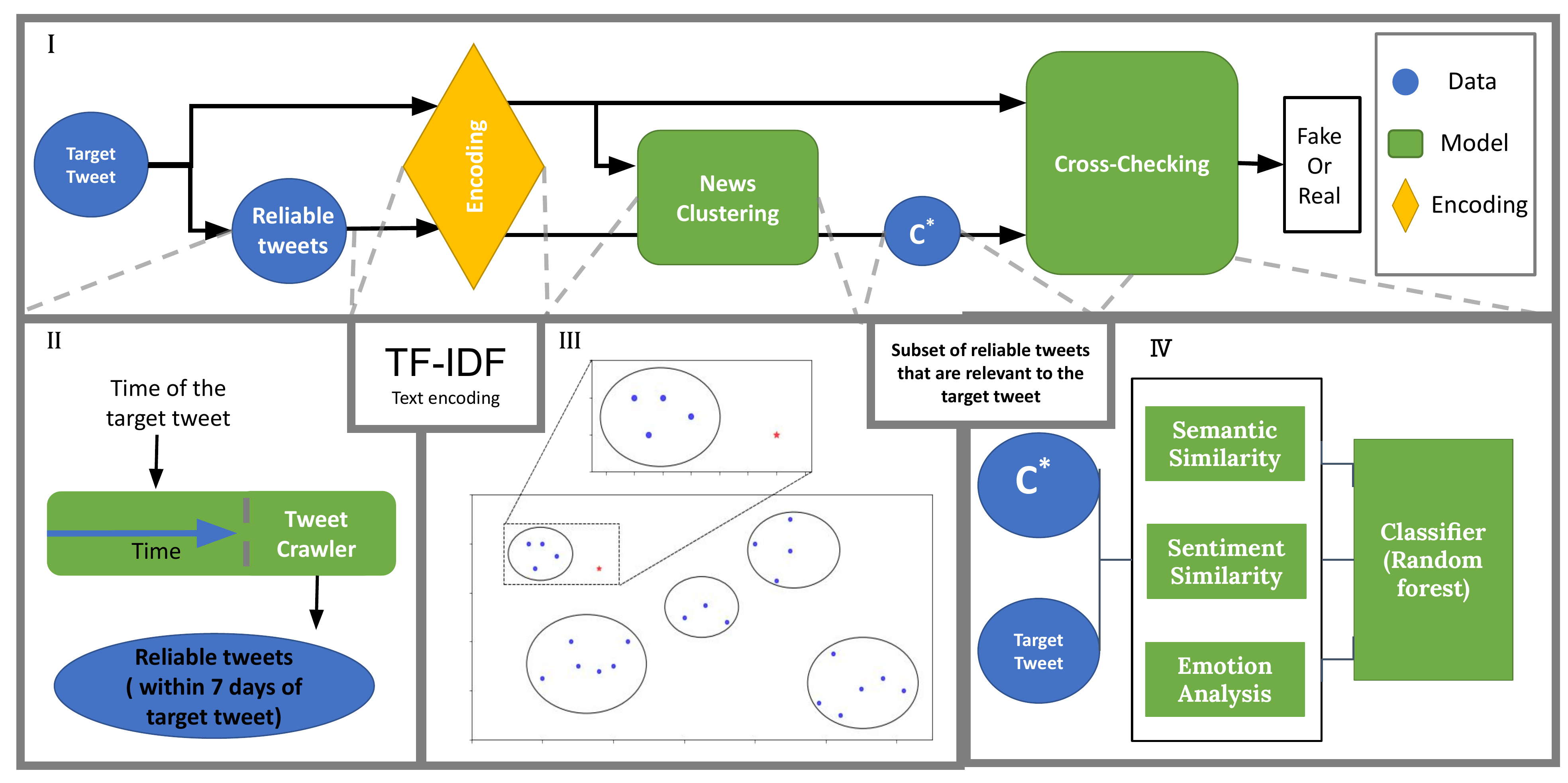}
\caption{This diagram shows the different parts of our model to process the tweets and find out their veracity. The first step is gathering data from \reliable{}. After encoding tweets' texts into vectors, we find the cluster of stories associated with the \targettweet{}, $\mathbb{C}^{*}$. The final step is to cross-check the \targettweet{} with reliable tweets in $\mathbb{C}^{*}$. The data is represented by blue circles, pre-processing by a yellow diamond, and green rectangles are the symbol of models.  }
\label{fig:flowchart}
\end{figure*}

Figure \ref{fig:flowchart}-\rom{1} shows our full pipeline for the classification of a \targettweet{} as \fakenews{} or \truenews{}. We first will briefly discuss the pipeline as illustrated in figure \ref{fig:flowchart}.

We assume that we are given a tweet, and our task is to check its veracity. This could be a new tweet that shows up on Twitter. We refer to this as the ``\targettweet{}.'' 
We also assume that we have access to reliable sources of information that we can use as reference and cross-check the \targettweet{} against these sources. We refer to these sources as ``\reliable{}.'' For our work, we use tweets from Twitter accounts of well-established news agencies, such as Reuters, BBC, Associated Press, etc., for the \reliable. 

We collect all tweets from \reliable{} in the three days leading to the \targettweet{} and the three days following it (figure \ref{fig:flowchart}-\rom{2}). For example, for the tweet ``No marines were killed in the Kabul airport attack, they were just injured.'' which was posted on 26th of August 2021, we collect tweets from news agencies in the time interval between 23rd of August 2021 and 29th of August 2021. We refer to this collection as `` $\mathcal{C}$'' of tweets. We need a mathematical representation of texts of tweets, for which we use TF-IDF encoding \cite{ullman2011mining} (figure \ref{fig:flowchart}).
\\

Set $\mathcal{C}$ comprises many different news stories, and not all of them are related to the \targettweet{}.
The next step is to cluster $\mathcal{C}$ into news stories. For this, we use a clustering algorithm that uses the text from the tweets and clusters them based on the similarity in the vocabulary used in the tweets. 
Mathematically, the clustering can be described as a function $f_c$ that assigns an index to any tweet. 
The result of the clustering is a set of clusters $\{\mathcal{C}_{i}\}$. Each $\mathcal{C}_{i}$ contains news on a specific news story. 
For instance, the time interval 23rd of August and 29th of August 2021 contains five clusters with stories about the Ida hurricane, a regional conference in Iraq, the Kabul airport attack, the Kabul airport evacuation, and the US FDA full approval of Pfizer/BioNTech’s COVID-19 vaccine.

Once we have the clusters, we need to identify the cluster to which the \targettweet{} belongs. For this, we use the same \clustering{}, i.e., $f_c$. Given the \targettweet{}, the clustering function finds the cluster that best matches the \targettweet{}. We refer to this as $\mathbb{C}^{*}$. 
For example, the \targettweet{} “No marines were killed in the Kabul airport attack, they were just injured” will be placed in the cluster of news about the Kabul airport attack. \\
Figure \ref{fig:flowchart}-\rom{3} schematically shows how tweets are clustered. 
Each point represents a tweet. The blue circles are tweets from \reliable{}, and the red star is the \targettweet{}. 
Circles represent clusters, i.e., news stories. By this point, we have identified the relevant news story from \reliable{}. 
The next step is to cross-check the content of the \targettweet{} against the reliable tweets in its corresponding cluster, $\mathbb{C}^{*}$. If the \targettweet{} is not aligned with $\mathbb{C}^{*}$, it is likely fake news. 
For example, if the \targettweet{} is ``No marines were killed in the Kabul airport attack, they were just injured.'', and in the corresponding cluster, we have a tweet like ``12 U.S. service members killed in terrorist attack outside Kabul airport, U.S. official says'' from The Washington Post, the \targettweet{} will be labeled as fake. 
This line of reasoning is inspired by how humans usually check the veracity of news. \\

We use a second machine learning model for cross-checking,  and we refer to it as the ``\crosscheck{}''. A schematic picture of this model is shown in figure \ref{fig:flowchart}-\rom{4}.
The \crosscheck{} relies on the semantic, sentiment similarities, and emotional characteristics between the \targettweet{} and tweets in $\mathbb{C}^*$ to determine if the \targettweet{} is aligned with the \reliable{}. We use separate models to characterize sentiment and semantic similarities, and emotional characteristics. 

The \crosscheck{} (figure \ref{fig:flowchart}-\rom{3}) is different from the \clustering{} (figure \ref{fig:flowchart}-\rom{4}) in the sense that for clustering, we put more emphasis on the words in the tweet, whereas for cross-checking, we train our model to pay more attention to the meaning. Different techniques can be used for cross-checking. These different techniques do not necessarily agree with each other. As an illustration, a \targettweet{} may have the same sentiment as the ones in \reliable{}, but with different semantics. The final piece of the pipeline is a model that takes the result of all the different comparison techniques (e.g., sentiment, semantic) and decides if the \targettweet{} is aligned with the relevant tweets from \reliable{} (i.e., $\mathbb{C}^{*}$). This model has to consider all the different comparison methods and weigh their results when they do not agree with each other. \\
We now describe each element of the pipeline in more detail. Specifically, in the following subsections, we will describe the model for clustering tweets based on their content, the models used for feature extraction from tweets, and the final classification based on different comparison methods.

\subsection{Clustering Model}

The first element is the clustering model. The \clustering{} takes the collection of the tweets from \reliable{} and clusters them to different groups that have similar news content.

For the clustering model, we need a mathematical representation of the text of the tweets in the collection. At this stage, we focus on the words used in tweets. For instance, if two tweets share words like ``Kabul airport'' and ``US military'', it is likely that they are about the same story. In NLP (Natural Language Processing), encodings like ``bag of words'' \cite{harris1954distributional, mctear2016conversational} or TF-IDF can provide representations based on the vocabularies used in the tweets. 

The bag-of-words encoding uses a corpus of words and represents any given text with a vector of the length of the corpus. If a word in the corpus is present in the text, the corresponding element of the vector would be the frequency of the word in the text. TF-IDF provides a similar encoding. The difference is that the elements of the vector are weighted based on the ``inverse document frequency (IDF)''. This reduces the weights of words that frequently occur, such as ``the''. On the other hand, it increases the weight of the words that rarely occur \cite{jones1972statistical}.

Typically, for TF-IDF, the corpus is a collection of all the words used in text samples (with some restrictions on their frequencies). This leads to a long and sparse representation. For our purpose, it is more meaningful to form a corpus of significant keywords. For instance, words like ``in'' and ``the'' are likely to appear in various news but carry little weight in discriminating those stories. In contrast, words like ``Kabul'', ``Taliban'', and ``attack'', are more expressive. We use named entities to form our corpus of words. Named entities are objects such as organizations, people,  Geopolitical Entity (GPE). We use a python library called ``SpaCy'' \cite{spacy} to identify the named entities \cite{choi2015depends}.

We use the collection of named entities as the corpus for our TF-IDF model to vectorize the text of a given tweet. Then we apply this encoding to the collection of tweets from \reliable{}, which gives a vector for each tweet from the collection.\\
Next, we apply the k-means clustering algorithm to the vectorized tweets. This gives clusters of tweets where each cluster contains tweets that have similar named entities. For the k-means algorithm, we need to identify the number of clusters before the clustering. However, the number of clusters depends on $\mathbb{C}^{*}$ and could change. We pick the number of clusters, k, such that the clustering maximizes the mean Silhouette Coefficient \cite{rousseeuw1987silhouettes}. The mean silhouette coefficient characterizes the ratio of intra-cluster distance and inter-cluster distance. Maximizing the mean silhouette coefficient indicates that clusters are well-apart compared to their internal distance. Finally, we have to assign the \targettweet{} to one of the clusters. To do so, we apply the same TF-IDF encoding to the \targettweet{} and vectorize the text of the \targettweet{}. We then apply the k-means model, trained on the collection, to the \targettweet{}. This returns the cluster that shares the largest number of named entities with the \targettweet{}. Finally, we use the cosine-similarity to filter the closest tweets inside that cluster for the \targettweet{}. Table \ref{table:reliables to compare with the example tweet} gives an example of a \targettweet{} and tweets in its corresponding cluster, $\mathbb{C}^{*}$. Note that clustering news based on the vocabulary is not perfect. For instance, the vocabularies in one tweet may match multiple clusters.
This is because, fundamentally, it is sometimes difficult to separate news stories, as there may be multiple stories that are related. This leads to overlapping clusters. One may consider soft clustering algorithms for this application \cite{bezdek2013pattern}.

\begin{table}[h!]
\centering
{\renewcommand{\arraystretch}{1.0}
\begin{tabular}{|>{\centering\arraybackslash}m{.2\linewidth}|>{\centering\arraybackslash}m{.8\linewidth}|}
\hline
\multicolumn{2}{|c|}{tweet examples} \\
\hline \centering
Twitter Account&Tweet Text\\
\hline \centering
CNN& ``At least 12 members of the US military were killed and at least 15 more were injured in the attack at Kabul's airport, Gen. Kenneth ``Frank'' McKenzie, head of the US Central Command, said at a briefing.''\\
\hline \centering
New York Times& ``Breaking News: At least 12 U.S. service members and scores of Afghan civilians were killed in the bombings outside the Kabul airport, officials said.''\\
\hline \centering
Associated Press& ``Two U.S, officials say at least 12 U.S. service members were killed in the Afghanistan bombings, including 11 Marines and one Navy medic. Officials say a number of U.S. troops were wounded.''\\
\hline \centering
BBC World& ``Twelve US service members - 11 marines and a Navy medic - were killed in Kabul airport attack, officials tell US media.''\\
\hline \centering
The \targettweet& ``No marines were killed in the Kabul airport attack, they were just injured.''\\
\hline
\end{tabular}
}
\caption{Example of $\mathbb{C}^{*}$ for the \targettweet{} of  ``No marines were killed in the Kabul airport attack, they were just injured''. By comparing the \targettweet{} with these tweets from reliable sources, we can see that this news is not \truenews. }
\label{table:reliables to compare with the example tweet}
\end{table}

\subsection{Cross-Checking Model}
At this point, we have the \targettweet{} as well as its corresponding story from \reliable{} (i.e., $\mathbb{C}^{*}$). The function of the \crosscheck{} is to compare the \targettweet{} with tweets in $\mathbb{C}^{*}$ and determine if the \targettweet{} is aligned with our \reliable{} or not. 

To train a model to classify the \targettweet{} as \truenews{} or \fakenews{}, we extract some features from tweets. Specifically, we use the semantic, sentiment, and emotion of a tweet to compare the \targettweet{} and \reliable{}. Figure \ref{fig:flowchart}-\rom{4} shows the diagram for our \crosscheck{}. In the following subsections, we will describe the NLP tools we use to extract each of these in detail. 

\subsubsection{Semantic Similarity}
We want to measure how semantically close the \targettweet{} is to tweets in $\mathbb{C}^{*}$. This time, we use word-embedding for text representation. A word-embedding can be seen as a mapping from the bag-of-words representation to an encoding that is semantically meaningful. Specifically, word-embedding assigns real-valued vectors to texts or words in a way that preserves their semantic relation. Different word-embedding techniques have been developed for NLP applications. Here, we use a technique known as ``FastText.'' We trained this word embedding on Common Crawl and Wikipedia \cite{grave2018learning}. For two semantically close tweets, their embedding vector should be relatively close. To find the distance between two tweets, we use the cosine between their corresponding vectors. If the two tweets are aligned, the cosine should be close to one. If they are distant, the cosine would be small. We use the cosine as a similarity measure  to compare the two vectors. This provides one of the three elements that will be used for checking the alignment of the \targettweet{} with respect to \reliable{}.

\subsubsection{Sentiment Similarity}
In addition to semantic, we examine the sentiment difference between the \targettweet{} and reliable tweets. For this, we train a semantic analyzer model using fastText over 3.6 million reviews from the Amazon dataset \cite{he2016ups,mcauley2015image}. Then we check the sentiment for the \targettweet{} and \reliable{}. We use  the difference between their sentiment values as a feature. For example, the sentiment of the \targettweet{} ``The India flood did not have any victims.'' is zero, i.e., neutral sentiment. The corresponding reliable tweet is ``At least 15 dead and dozens missing in India floods '', which has a sentiment of $-1$, i.e., negative. The difference of the sentiment values is $-1$. This means that for this case, we can identify the \targettweet{} as fake using its sentiment. \\
However, cross-checking is not always this simple, and we can not always rely on comparing sentiments to cross-check the veracity of the \targettweet{}. 
It is possible to make fake news with the same sentiment as \reliable{}. Therefore, for our \crosscheck{}, we also analyze the emotional content of the \targettweet{}.

\subsubsection{Emotion Analysis}
Humans can identify emotions from the text. Emotions that can be received from \fakenews{} news might differ from \truenews{} news. We expect \fakenews{} news to cause emotions such as fear and surprise. We use a tool called text2emotion to extract emotions from the text of the tweets. Text2emotion is a python library that, given an input text, assigns one of the Happy, Angry, Sad, Surprise, or Fear emotions to it. For example, the \targettweet{} ``No marines were killed in the Kabul airport attack, they were just injured'' stimulates both sadness and happiness, while reliable tweets from the same cluster cause fear and anger. 

\subsubsection{Data and Classification}
\begin{figure}[h!]
\centering
\includegraphics[scale=0.4]{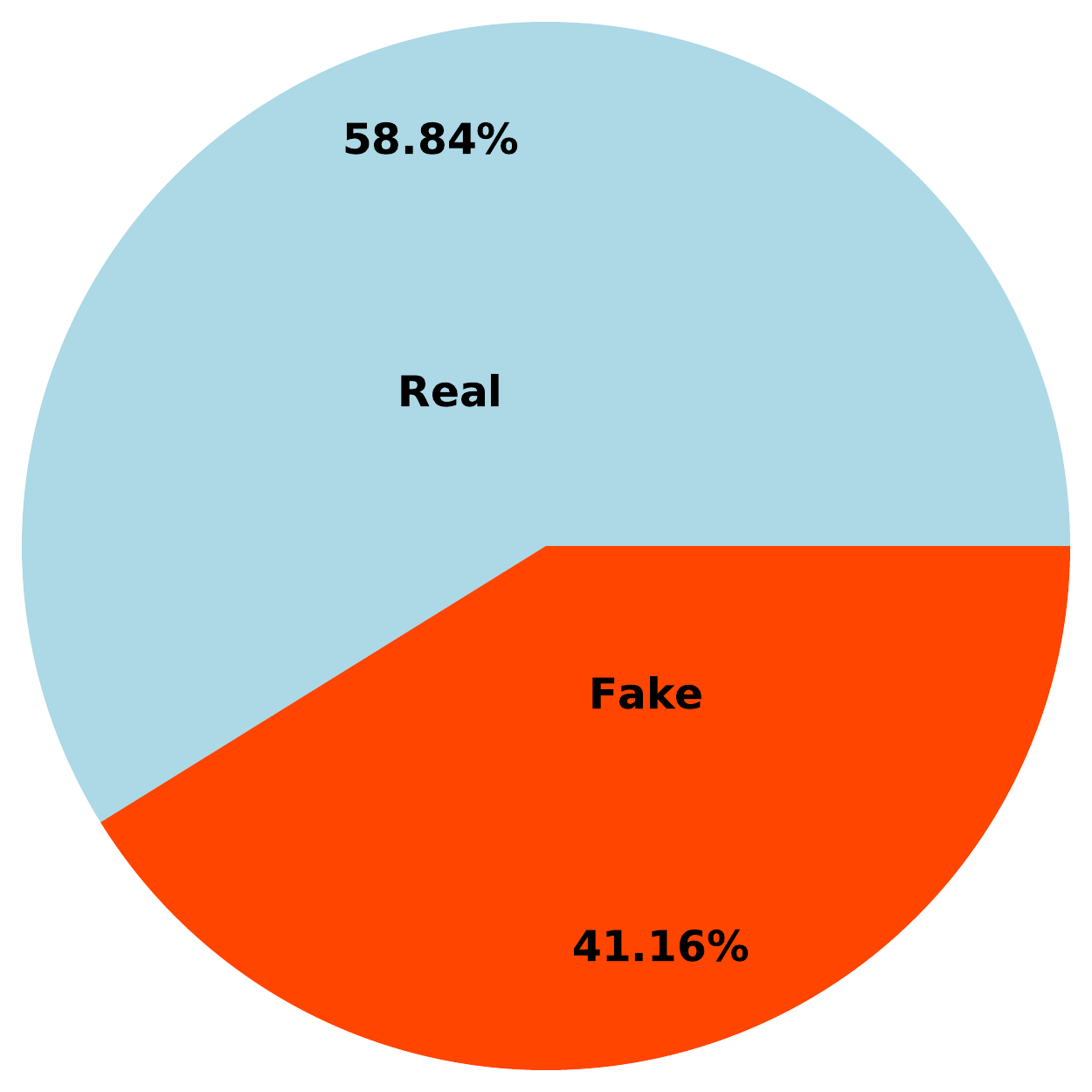}
\caption{Distribution of classes in our Fake-Real dataset. Our dataset has 1921 fake and 2764 real tweets}
\label{fig:pie_dist_data}
\end{figure}

The extracted features represent the semantic and sentiment differences and emotional characteristics between the \targettweet{} and the tweets from \reliable{}. For the last part of our pipeline, we need a model that takes these extracted features as inputs and classifies if the \targettweet{} is \truenews{} or \fakenews{}.\\ 
For this model, we need a labeled dataset. We used the ``Fake and Real News dataset'' published on Kaggle \footnote{https://www.kaggle.com/clmentbisaillon/fake-and-real-news-dataset}. This dataset is based on Ahmed et al. \cite{ahmed2018detecting}. It includes 9365 \fakenews{} news articles and 18984 \truenews{} articles in English. The Fake and Real News dataset contains article text, article type, article label (\fakenews{} or \truenews{}), article title, and article date. These articles are longer than typical tweets. Therefore, we use article titles to test and train our model. These article titles act similar to tweets. We refer to this dataset as the ``Fake-Real'' dataset. We use the Fake-Real dataset to train and test our Cross-checking model and identify tweets' veracity. Figure \ref{fig:pie_dist_data} shows the distribution of classes in our dataset.\\ 

\begin{figure*}
\centering
\includegraphics[scale=0.37]{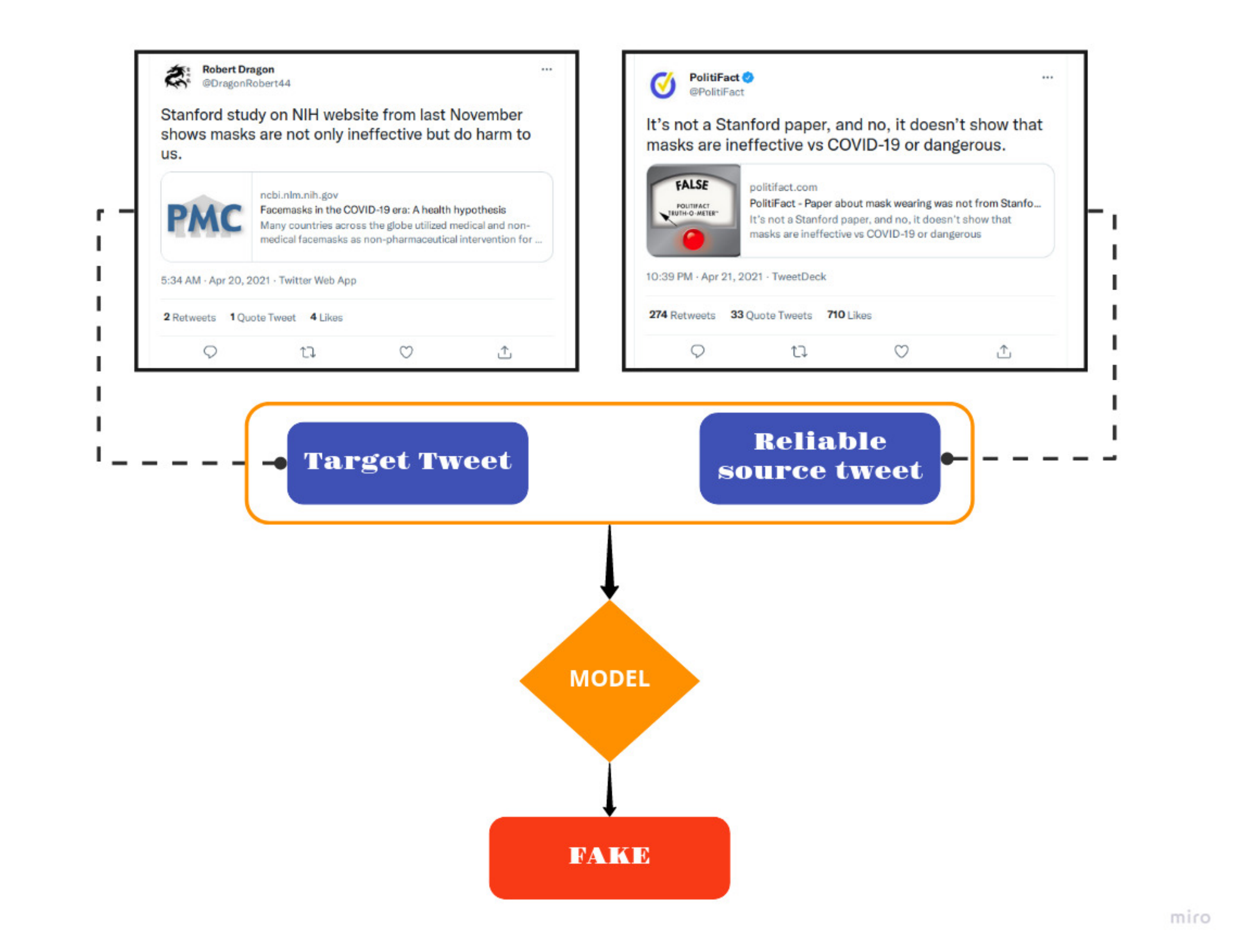}
\caption{An example Tweet and the corresponding classification process. On the left side, the tweet contains \fakenews{} news, and the right side contains a tweet from the PolitiFact Twitter account. The model cross-checks the \targettweet{} with a tweet from a reliable source to find out its veracity.}
\label{fig:dataset_schema}
\end{figure*}

We train a random forest model on this dataset. We use the extracted features (i.e., semantic and sentimental differences and emotional characteristics) of each tweet and its label to train the random forest model. Figure \ref{fig:dataset_schema} includes an example of a tweet containing \fakenews{} news and a tweet from a reliable source, which is not aligned with the \targettweet{}.

\section{Results}
\begin{figure}
\centering
\includegraphics[width = \columnwidth]{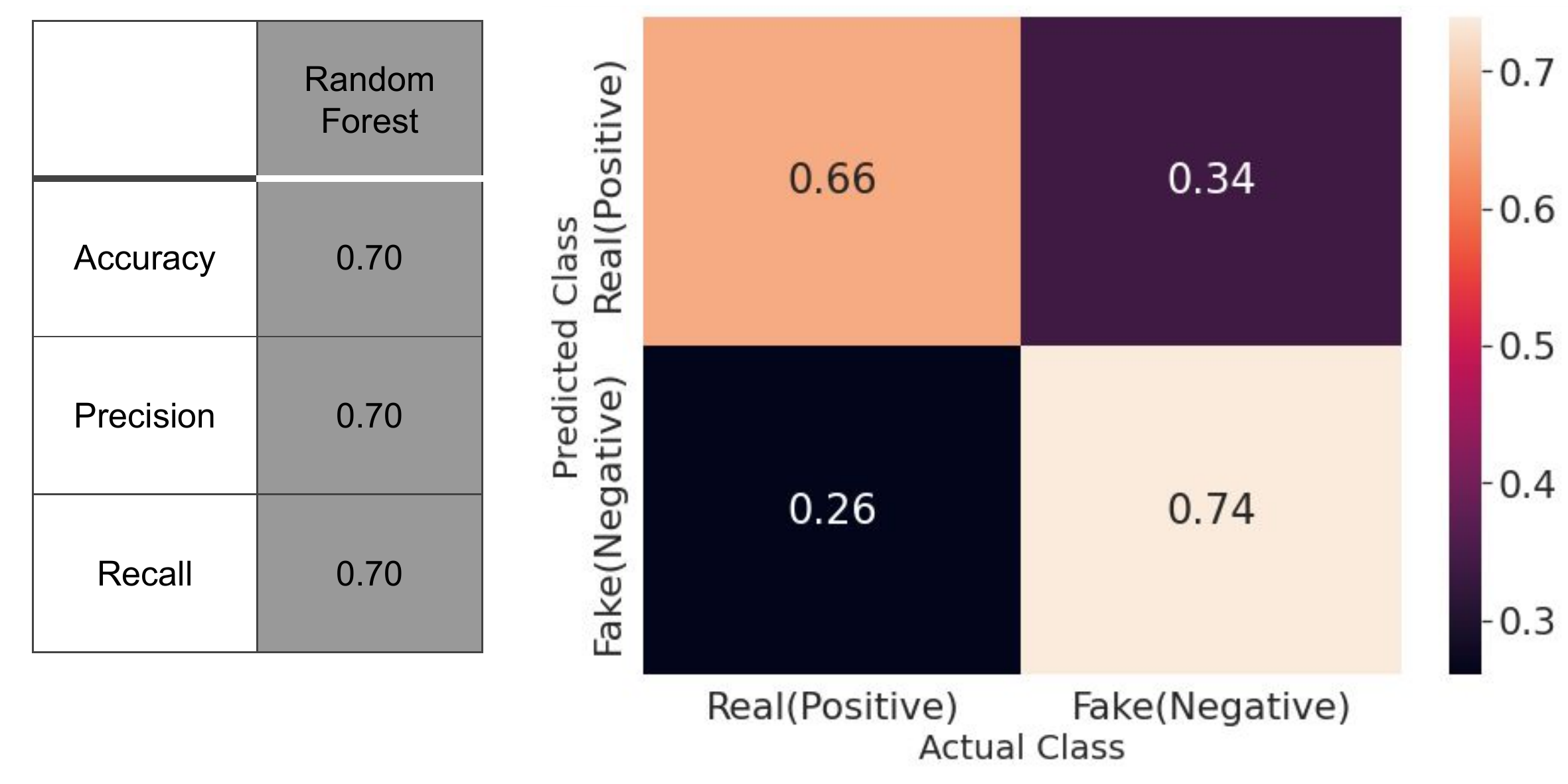}
\caption{The results for detection of \fakenews{} tweets. Random Forest gives an accuracy of 0.70}
\label{fig:diagram}
\end{figure}

We use three metrics to evaluate the performance  of our classifiers. These metrics are accuracy, precision, and recall \footnote{https://en.wikipedia.org/wiki/Precision\_and\_recall}. Figure \ref{fig:diagram} shows these metrics for our model. Our random forest model has an accuracy of  0.70, a precision of 0.70, and a recall of 0.70 for the class \fakenews{} of tweets. We balanced the population of fake and real tweets in our training dataset for these results.

\begin{figure}
\centering
\includegraphics[width = \columnwidth]{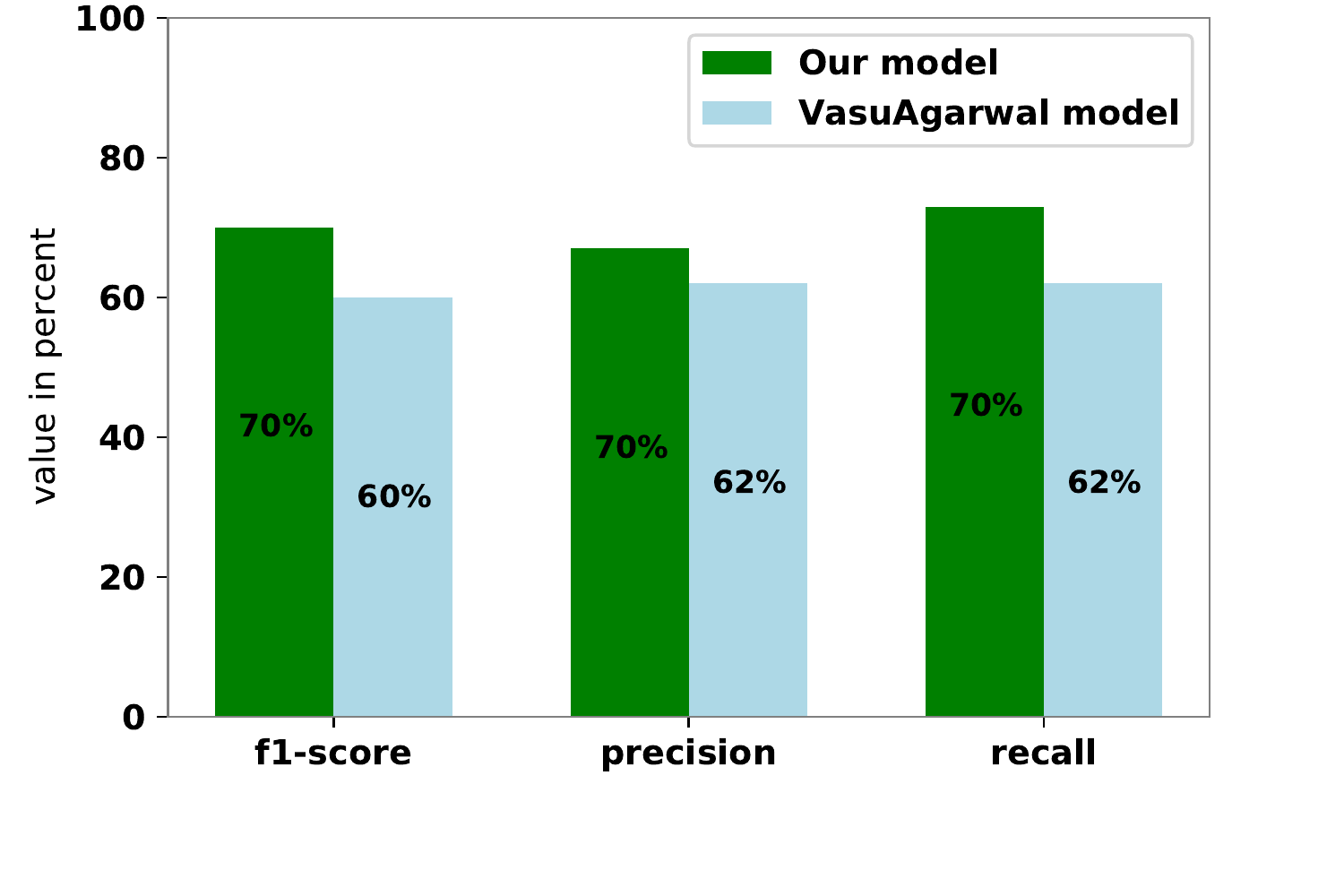}
\caption{Comparison of the Random Forest results with \cite{agarwal2019analysis}. 
The comparisons were made for three different metrics: f1-score, precision, and recall. For all metrics, our model outperforms the result from \cite{agarwal2019analysis}. In all three metrics, our model shows better results. 
}
\label{fig:compare_bar_chart_bw_2}
\end{figure}

We compared our result with \cite{agarwal2019analysis}, which we refer to as the VasuAgarwal model, in Figure \ref{fig:compare_bar_chart_bw_2}. As was explained in the Introduction, this work is comparable to our model in two major senses. First, they only use text for the identification of \fakenews{} tweets. Second, similar to our model, their work is not limited to a specific story and can be used for any given target tweet. Figure \ref{fig:compare_bar_chart_bw_2} shows that our model achieves better performance for all metrics \cite{agarwal2019analysis}.


We also tested our model to determine the veracity of tweets from some famous politicians. Table \ref{table:tweet_examples} shows some examples, and our model predicts these tweets' veracity correctly.\\
\begin{table}[h!]
\centering
{\renewcommand{\arraystretch}{1.5}
\begin{tabular}{|>{\centering\arraybackslash}m{.15\linewidth}|>{\centering\arraybackslash}m{.44\linewidth}|>{\centering\arraybackslash}m{.18\linewidth}|>{\centering\arraybackslash}m{.18\linewidth}|}
\hline \centering
Tweeter&Tweet Text & Prediction & PolitiFact \\
\hline \centering
Donald Trump&``Michigan just refused to certify the election results'' & \fakenews{} & \fakenews{}\\
\hline \centering
Bernie Sanders&``You know what Amazon paid in federal income taxes last year? zero'' & \truenews{} & \truenews{}\\
\hline
\end{tabular}
}
\caption{Example of tweets containing \fakenews{} and \truenews{} news from famous politicians. These tweets, labeled as \fakenews{} or \truenews{} by PolitiFact, have been correctly classified by our model. }
\label{table:tweet_examples}
\end{table}
There exist some tweets in the dataset that were mislabeled. 
Interestingly, our model managed to catch these mistakes. 
Table \ref{table:wrong_right_predicted_table} shows an example that was labeled as a \fakenews{} tweet, whereas it is actually \truenews{}. Our model labels it as a \truenews{} tweet. This indicates that the performance of our model is probably even higher than the numbers in figure \ref{fig:diagram}. As a result, we expect to get a higher performance for our model on a more accurate dataset.
\begin{table}[h!]
\centering
{\renewcommand{\arraystretch}{1.5}
\begin{tabular}{|>{\centering\arraybackslash}m{.45\linewidth}|>{\centering\arraybackslash}m{.20\linewidth}|>{\centering\arraybackslash}m{.15\linewidth}|>{\centering\arraybackslash}m{.15\linewidth}|}

\hline \centering
Tweet Text & Prediction (our model) & Reality & Dataset Label  \\
\hline \centering
``Ben Jerry’s ice cream founders were arrested at the US capital''&\truenews{}&\truenews{}&\fakenews{}\\
\hline
\end{tabular}
}
\caption{Example of a tweet that was mislabeled in the dataset and our model classified  correctly.}
\label{table:wrong_right_predicted_table}
\end{table}
\newline

\section{Discussions and Conclusion}

We built a model for fake news detection, which is inspired by human behavior. Namely, our model compares tweets with \reliable{} to find whether they contain \fakenews{} or \truenews{} news. We wanted to build this model in a way that is not limited to viral stories. Therefore, we trained our model on a dataset containing diverse news stories. For this case, other famous datasets exist, such as Liar \cite{wang2017liar}. Although the Liar dataset is more widely used, we chose the Fake and Real News dataset since it is larger and more diverse.\\

There are a few assumptions and details about the work that we need to emphasize. First, in order to compare tweets with \reliable{}, we collect tweets from Twitter accounts of reliable news agencies. Here we used BBC, Deutsche Welle, The New York Times, Al Jazeera, Associated Press, The Washington Post, etc., for the reliable sources. Choosing more than one reliable source not only helps us to cover more tweets, but also helps us determine its veracity more accurately. Apart from that, based on the subject of news, whether political, scientific, economic, etc., we must choose \reliable{} related to that subject.\\ 
Note that in some cases, we might not be able to assign a cluster to the \targettweet. For instance, there may be \fakenews{} news that has no reflection in the media. In this case, we will be unable to find reliable news for that specific \targettweet{}. 

Also, note that the choice of \reliable{} can significantly influence the classification result. For instance, using Fox News as a reliable source significantly changes the results. This is because fake news detection is, to some extent, subjective. Even in real life, different people may not agree on whether news is fake or real based on what they use to cross-check the news against. We tried to pick more factual news outlets based on some research\footnote{https://www.adfontesmedia.com/static-mbc/} for our choice here. Nevertheless, we understand that not everyone may agree with the choices of \reliable{}. However, the approach can be adapted to what the user or the target audience trust for reliable information. 

Another point that we need to emphasize is that we choose the titles of news articles from the ``Fake and Real news dataset'' as our \targettweet{}s. These titles are similar to tweets because they are short texts containing news. 
We also use these short texts to extract tweets from Twitter and determine the veracity of extracted tweets using these titles. 
We believe that our model can be used for long text as well. In that case, we can use articles from reliable news agencies to compare a long target text with them. In the case of long texts, the three features we extract might not be enough. This is because the overlap between news articles is more in long texts, and we probably will need other features for the cross-checking. 

Thirdly, an important aspect of our work is that our model is not limited to a specific story. Often models that are developed to use news content for fake news detection focus on specific stories on social media \cite{qazvinian2011rumor}, and their applications are limited to those specific stories. Our model is not limited to specific stories and can cover a wider range of tweets. 
Also, note that, as expected, compared to story-focused models, our approach is  less accurate. However, it outperforms other approaches that are not limited to specific stories. One of the main bottlenecks of our approach is news clustering. Our news clustering is based on vocabulary and is not perfect. For instance, the vocabularies in one tweet may match multiple clusters.
This is because, fundamentally, it is sometimes difficult to separate news stories, as there may be multiple stories that are related. This leads to overlapping clusters. One may consider soft clustering algorithms for this application \cite{bezdek2013pattern}.

Another key point about our approach is that for a given \targettweet{}, we collect tweets from \reliable{} from both before and after the \targettweet{} is posted. 
First, note that we can easily change this and limit it to tweets before the \targettweet{}. We think it is important since news agencies might have delayed broadcasting some news for any reason. The last remark is that we did not put much effort into fine-tuning our model. We expect that our result could significantly improve with fine-tuning the hyper-parameters of the model.
In summary, we presented a model to predict the veracity of news on Twitter by checking the deviations from reliable sources. The results outperform similar works. Our model can be divided into four parts. We first build a collection of tweets from reliable sources from the same week as the \targettweet{}. Next, we cluster them based on their stories and identify the cluster related to the \targettweet{}. Last, we build a model that cross-checks the \targettweet{} with reliable tweets based on their semantic and sentimental differences and emotional characteristics.
There is room to improve the model and results by optimizing different model parts and using a better dataset for training. The high efficiency of the model shows that the model inspired by human behavior can become biased, too, since the main idea of this approach is based on the behavior of checking the reliable source. As discussed, this can become toxic too. 

\appendix
\section{Pre-processing}
We run some pre-processing on the data that we collect from Twitter. 
We remove stop words, punctuation, and lowercase words in tweets. Removing this irrelevant information reduces the size of data. Since this information does not contribute towards the actual meaning of a sentence, discarding them will improve the performance.
In addition, we remove Twitter handles, email addresses, and URLs. In the end, since a tweet might contain non-English characters, we fixed any encoding error in the text using Beautiful-Soup \cite{richardson2007beautiful}. The Beautiful-Soup module automatically detects the encoding method of the text and converts it to a suitable format.

\bibliographystyle{aipauth4-1}
\bibliography{references}

\begin{thebibliography}{10}

\bibitem{agarwal2019analysis}
Vasu Agarwal, H~Parveen Sultana, Srijan Malhotra, and Amitrajit Sarkar.
\newblock Analysis of classifiers for fake news detection.
\newblock {\em Procedia Computer Science}, 165:377--383, 2019.

\bibitem{ahmed2018detecting}
Hadeer Ahmed, Issa Traore, and Sherif Saad.
\newblock Detecting opinion spams and fake news using text classification.
\newblock {\em Security and Privacy}, 1(1):e9, 2018.

\bibitem{allcott2017social}
Hunt Allcott and Matthew Gentzkow.
\newblock Social media and fake news in the 2016 election.
\newblock {\em Journal of economic perspectives}, 31(2):211--36, 2017.

\bibitem{balmas}
M~Balmas.
\newblock When fake news becomes real: Combined exposure to multiple news
  sources and political attitudes of inefficacy, alienation, and cynicism.
\newblock {\em Communication Research}, 41(2):430–454, 2014.

\bibitem{bezdek2013pattern}
James~C Bezdek.
\newblock {\em Pattern recognition with fuzzy objective function algorithms}.
\newblock Springer Science \& Business Media, 2013.

\bibitem{buntain2017automatically}
Cody Buntain and Jennifer Golbeck.
\newblock Automatically identifying fake news in popular twitter threads.
\newblock In {\em 2017 IEEE International Conference on Smart Cloud
  (SmartCloud)}, pages 208--215. IEEE, 2017.

\bibitem{choi2015depends}
Jinho~D Choi, Joel Tetreault, and Amanda Stent.
\newblock It depends: Dependency parser comparison using a web-based evaluation
  tool.
\newblock In {\em Proceedings of the 53rd Annual Meeting of the Association for
  Computational Linguistics and the 7th International Joint Conference on
  Natural Language Processing (Volume 1: Long Papers)}, pages 387--396, 2015.

\bibitem{ecker2017reminders}
Ullrich~KH Ecker, Joshua~L Hogan, and Stephan Lewandowsky.
\newblock Reminders and repetition of misinformation: Helping or hindering its
  retraction?
\newblock {\em Journal of Applied Research in Memory and Cognition},
  6(2):185--192, 2017.

\bibitem{grave2018learning}
Edouard Grave, Piotr Bojanowski, Prakhar Gupta, Armand Joulin, and Tomas
  Mikolov.
\newblock Learning word vectors for 157 languages.
\newblock In {\em Proceedings of the International Conference on Language
  Resources and Evaluation (LREC 2018)}, 2018.

\bibitem{harris1954distributional}
Zellig~S Harris.
\newblock Distributional structure.
\newblock {\em Word}, 10(2-3):146--162, 1954.

\bibitem{he2016ups}
Ruining He and Julian McAuley.
\newblock Ups and downs: Modeling the visual evolution of fashion trends with
  one-class collaborative filtering.
\newblock In {\em proceedings of the 25th international conference on world
  wide web}, pages 507--517, 2016.

\bibitem{helmstetter2018weakly}
Stefan Helmstetter and Heiko Paulheim.
\newblock Weakly supervised learning for fake news detection on twitter.
\newblock In {\em 2018 IEEE/ACM International Conference on Advances in Social
  Networks Analysis and Mining (ASONAM)}, pages 274--277. IEEE, 2018.

\bibitem{spacy}
Matthew Honnibal, Ines Montani, Sofie Van~Landeghem, and Adriane Boyd.
\newblock {spaCy: Industrial-strength Natural Language Processing in Python},
  2020.

\bibitem{jones1972statistical}
Karen~Sparck Jones.
\newblock A statistical interpretation of term specificity and its application
  in retrieval.
\newblock {\em Journal of documentation}, 1972.

\bibitem{kaminska2017module}
I~Kaminska.
\newblock A module in fake news from the info-wars of ancient rome.
\newblock {\em Financial Times. Accessed}, 28:03--18, 2017.

\bibitem{kwon2013prominent}
Sejeong Kwon, Meeyoung Cha, Kyomin Jung, Wei Chen, and Yajun Wang.
\newblock Prominent features of rumor propagation in online social media.
\newblock In {\em 2013 IEEE 13th international conference on data mining},
  pages 1103--1108. IEEE, 2013.

\bibitem{matsa2018news}
Katerina~Eva Matsa and Elisa Shearer.
\newblock News use across social media platforms 2018.
\newblock {\em Pew Research Center}, 10, 2018.

\bibitem{mcauley2015image}
Julian McAuley, Christopher Targett, Qinfeng Shi, and Anton Van Den~Hengel.
\newblock Image-based recommendations on styles and substitutes.
\newblock In {\em Proceedings of the 38th international ACM SIGIR conference on
  research and development in information retrieval}, pages 43--52, 2015.

\bibitem{mctear2016conversational}
Michael~Frederick McTear, Zoraida Callejas, and David Griol.
\newblock {\em The conversational interface}, volume~6.
\newblock Springer, 2016.

\bibitem{naeem2021exploration}
Salman~Bin Naeem, Rubina Bhatti, and Aqsa Khan.
\newblock An exploration of how fake news is taking over social media and
  putting public health at risk.
\newblock {\em Health Information \& Libraries Journal}, 38(2):143--149, 2021.

\bibitem{pavleska2018performance}
Tanja Pavleska, Andrej {\v{S}}kolkay, Bissera Zankova, Nelson Ribeiro, and Anja
  Bechmann.
\newblock Performance analysis of fact-checking organizations and initiatives
  in europe: a critical overview of online platforms fighting fake news.
\newblock {\em Social media and convergence}, 29, 2018.

\bibitem{qazvinian2011rumor}
Vahed Qazvinian, Emily Rosengren, Dragomir Radev, and Qiaozhu Mei.
\newblock Rumor has it: Identifying misinformation in microblogs.
\newblock In {\em Proceedings of the 2011 Conference on Empirical Methods in
  Natural Language Processing}, pages 1589--1599, 2011.

\bibitem{reuter2015online}
Ora~John Reuter and David Szakonyi.
\newblock Online social media and political awareness in authoritarian regimes.
\newblock {\em British Journal of Political Science}, 45(1):29--51, 2015.

\bibitem{richardson2007beautiful}
Leonard Richardson.
\newblock Beautiful soup documentation.
\newblock {\em Dosegljivo: https://www. crummy.
  com/software/BeautifulSoup/bs4/doc/.[Dostopano: 7. 7. 2018]}, 2007.

\bibitem{rousseeuw1987silhouettes}
Peter~J Rousseeuw.
\newblock Silhouettes: a graphical aid to the interpretation and validation of
  cluster analysis.
\newblock {\em Journal of computational and applied mathematics}, 20:53--65,
  1987.

\bibitem{shao2017spread}
Chengcheng Shao, Giovanni~Luca Ciampaglia, Onur Varol, Alessandro Flammini, and
  Filippo Menczer.
\newblock The spread of fake news by social bots.
\newblock {\em arXiv preprint arXiv:1707.07592}, 96:104, 2017.

\bibitem{shearer2021news}
Elisa Shearer and Amy Mitchell.
\newblock News use across social media platforms in 2020.
\newblock 2021.

\bibitem{ullman2011mining}
Jeffrey Ullman.
\newblock {\em Mining of massive datasets}.
\newblock Cambridge University Press, 2011.

\bibitem{vosoughi2018spread}
Soroush Vosoughi, Deb Roy, and Sinan Aral.
\newblock The spread of true and false news online.
\newblock {\em Science}, 359(6380):1146--1151, 2018.

\bibitem{wang2017liar}
William~Yang Wang.
\newblock " liar, liar pants on fire": A new benchmark dataset for fake news
  detection.
\newblock {\em arXiv preprint arXiv:1705.00648}, 2017.

\bibitem{xiong2019hashtag}
Ying Xiong, Moonhee Cho, and Brandon Boatwright.
\newblock Hashtag activism and message frames among social movement
  organizations: Semantic network analysis and thematic analysis of twitter
  during the\# metoo movement.
\newblock {\em Public relations review}, 45(1):10--23, 2019.

\bibitem{zhao2020fake}
Zilong Zhao, Jichang Zhao, Yukie Sano, Orr Levy, Hideki Takayasu, Misako
  Takayasu, Daqing Li, Junjie Wu, and Shlomo Havlin.
\newblock Fake news propagates differently from real news even at early stages
  of spreading.
\newblock {\em EPJ Data Science}, 9(1):1--14, 2020.

\end{thebibliography}

\end{document}